%% file: main.tex
\documentclass[twocolumn, switch]{article} %

\usepackage{preprint}

\usepackage{amsmath, amsthm, amssymb, amsfonts}

\usepackage[square, comma, numbers,sort&compress]{natbib}
\usepackage[utf8]{inputenc}	%
\usepackage[T1]{fontenc}	%
\usepackage{xcolor}		%
\usepackage[colorlinks = true,
            linkcolor = purple,
            urlcolor  = blue,
            citecolor = cyan,
            anchorcolor = black,
            backref = page]{hyperref}	%
\usepackage{booktabs} 		%
\usepackage{nicefrac}		%
\usepackage{microtype}		%
\usepackage{lineno}		%
\usepackage{float}			%

\usepackage{lipsum}		%

\usepackage{newfloat}
\DeclareFloatingEnvironment[name={Supplementary Figure}]{suppfigure}
\usepackage{sidecap}
\sidecaptionvpos{figure}{c}

\usepackage{titlesec}
\titlespacing\section{0pt}{12pt plus 3pt minus 3pt}{1pt plus 1pt minus 1pt}
\titlespacing\subsection{0pt}{10pt plus 3pt minus 3pt}{1pt plus 1pt minus 1pt}
\titlespacing\subsubsection{0pt}{8pt plus 3pt minus 3pt}{1pt plus 1pt minus 1pt}

\usepackage{tikz,xcolor,hyperref}

\definecolor{lime}{HTML}{A6CE39}
\DeclareRobustCommand{\orcidicon}{
	\begin{tikzpicture}
	\draw[lime, fill=lime] (0,0) 
	circle [radius=0.16] 
	node[white] {{\fontfamily{qag}\selectfont \tiny ID}};
	\draw[white, fill=white] (-0.0625,0.095) 
	circle [radius=0.007];
	\end{tikzpicture}
	\hspace{-2mm}
}
\foreach \x in {A, ..., Z}{\expandafter\xdef\csname orcid\x\endcsname{\noexpand\href{https://orcid.org/\csname orcidauthor\x\endcsname}
			{\noexpand\orcidicon}}
}

\title{Spatial-Temporal Multi-Cuts for \\Online Multiple-Camera Vehicle Tracking}

\usepackage{authblk}

\author{Fabian Herzog}
\author{Johannes Gilg}
\author{Philipp Wolters}
\author{Torben Teepe}
\author{Gerhard Rigoll}

\affil{Technical University of Munich\\\vspace{0.2em} \url{https://github.com/fubel/stmc}}

\usepackage{algorithm}
\usepackage{algpseudocodex}
\usepackage{booktabs}
\usepackage[caption=false]{subfig}
\usepackage{amssymb}
\usepackage{amsmath}
\usepackage{cleveref}
\usepackage{amsfonts}
\usepackage{multirow}

\newcommand{\ie}{\textit{i.e.}}
\newcommand{\eg}{\textit{e.g.}}
\newcommand{\etal}{et al. }

\DeclareMathOperator{\cycles}{cycles}

\DeclareMathOperator*{\nanmean}{nan\,mean}
\DeclareMathOperator*{\EMA}{EMA}

\usepackage[inline]{enumitem}

\usepackage{pifont}

\definecolor{cpdarkblue}{HTML}{045275}
\definecolor{cpgreen}{HTML}{00CD6C}
\definecolor{cpred}{HTML}{FF1F5B}

\renewcommand{\algorithmicrequire}{\textbf{Input:}}
\renewcommand{\algorithmicensure}{\textbf{Output:}}

\newcommand{\cmark}{{\color{cpgreen} \ding{51}}}

\newcommand{\cxmark}{{\color{cpdarkblue} \ding{82}}}%
\newcommand{\xmark}{{\color{cpred} \ding{55}}}%

\newcommand{\feat}{\operatorname{feat}}
\newcommand{\pos}{\operatorname{pos}}

\newcommand{\ours}{Ours}
\usepackage{bm}
\usepackage{optidef}

\newcommand{\bftab}{\fontseries{b}\selectfont}

\begin{document}

\twocolumn[ %
  \begin{@twocolumnfalse} %
  
\maketitle

\begin{abstract}
Accurate online multiple-camera vehicle tracking is essential for intelligent transportation systems, autonomous driving, and smart city applications. Like single-camera multiple-object tracking, it is commonly formulated as a graph problem of tracking-by-detection. Within this framework, existing online methods usually consist of two-stage procedures that cluster temporally first, then spatially, or vice versa. This is computationally expensive and prone to error accumulation.
We introduce a graph representation that allows spatial-temporal clustering in a single, combined step: New detections are spatially and temporally connected with existing clusters. By keeping sparse appearance and positional cues of all detections in a cluster, our method can compare clusters based on the strongest available evidence. The final tracks are obtained online using a simple multi-cut assignment procedure. Our method does not require any training on the target scene, pre-extraction of single-camera tracks, or additional annotations. 
Notably, we outperform the online state-of-the-art on the CityFlow dataset in terms of IDF1 by more than 14\%, and on the Synthehicle dataset by more than 25\%, respectively. The code is publicly available.
\end{abstract}
\vspace{1cm}

  \end{@twocolumnfalse} %
] %

\input{content/01_introduction}
\input{content/02_related}
\input{content/03_method}

\input{content/04_results}
\input{content/05_conclusion}

\normalsize
\bibliography{references}

\end{document}

%% file: content/01_introduction.tex
\begin{figure*}[ht!]
    \centering
    \includegraphics[width=\textwidth]{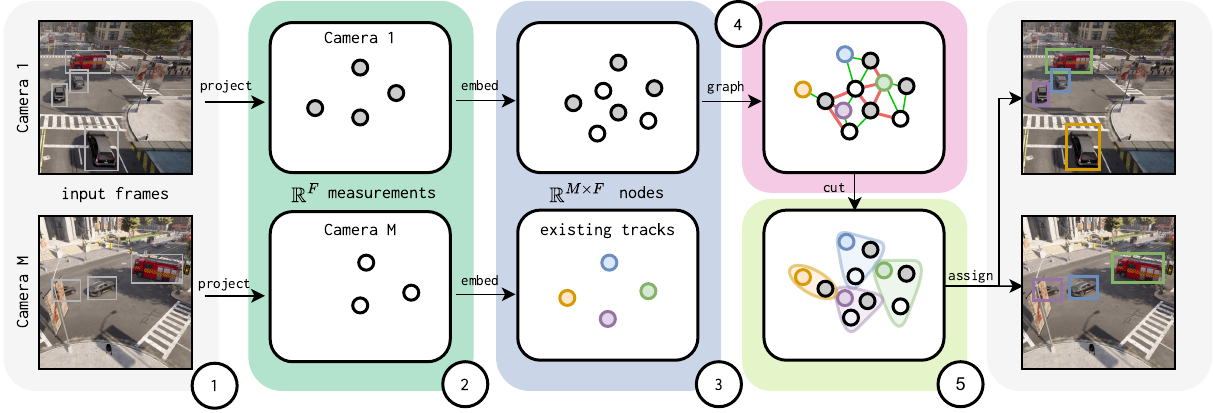}
    \caption{The proposed spatial-temporal multicuts (STMC). (1) Given the current video frames of all cameras, we apply a detector and extract the appearance features. (2) Features and detections are embedded in a common feature space. (3) Next, the measurements are aggregated into a \emph{superbox}-structure and connected to existing tracks. (4) Weights are computed on the basis of appearance and distance similarity. (5) The multicut solver then yields cluster candidates, and detections and tracks within clusters are assigned to each other in an assignment procedure.}
    \label{fig:teaser}
\end{figure*}
\section{Introduction}
\label{sec:introduction}

Multiple-target multiple-camera tracking (MTMCT) detects, localizes, and unambiguously identifies objects of interest on multiple cameras and over time. It shares some challenges with single-camera object tracking, such as dealing with occluded objects, problematic lighting and weather conditions, or target objects' varying size and look~\cite{dendorfer2019cvpr19,dendorfer2021motchallenge}. However, there are some critical and complex differences: On the one hand, introducing multiple cameras enables trackers to better estimate the underlying 3D geometry of a scene, allowing for more precise localization and resolution of identity switches~\cite{hou2020multiview,teepe2023earlybird}. On the other hand, it increases the amount of data processing needed and adds to the combinatorial complexity of the assignment problem.
Depending on the application, MTMC trackers use different strategies to solve these issues. Most methods follow the paradigm of tracking-by-detection~\cite{ristanicvpr2018, hou2019locality, hsu2019multi, luna2022graph, luna2021online, quach2021dyglip, shim2023fast, xu2017cross, he2020multi,  cheng2023rest, nguyen2022lmgp}. Often designed as a multi-stage approach, they first apply a detector and appearance feature extractor to construct measurements for each camera and time step. A subsequent process merges these measurements into consistent tracks. Within this framework, some methods first apply a single-camera tracker on each camera and then cluster the resulting tracks, either online or offline~\cite{he2019multi, qian2020electricity, he2020multi, luna2022graph, shim2023fast}. Other recent methods focus on fusing detections early in the spatial domain and reducing the tracking problem to the ground plane~\cite{teepe2023earlybird, Teepe_2024_CVPR}.

There are certain advantages and disadvantages to each of these paradigms:
Working on pre-extracted single-camera tracks enables the utilization of strong appearance cues. It also delegates some motion modeling to an external module, but errors made in the single-camera stage are hard to fix~\cite{specker2022improving}. Many online methods that work directly on detections or partial single-camera tracks accumulate errors due to their hierarchical structure or are specifically designed for the target dataset, for example, by introducing roadside annotations for vehicle tracking or manual time synchronization of cameras~\cite{liu2021city,shim2023fast}.
Early-fusion methods offer exact target localization and remove some complexity from the assignment problem but do not generalize well to unseen camera topologies~\cite{teepe2023earlybird,Teepe_2024_CVPR}. Moreover, there is limited availability of annotated data that is needed for better generalization~\cite{herzog2023synthehicle}.

We suggest a different approach that combines the advantages of multiple paradigms. By introducing a common graph representation for existing multiple-camera spatial-temporal tracks and single-camera detections, we solve the single- and multiple-camera data association problem in one single combined step. In this way, single-camera information helps multiple-camera associations, and multiple-camera information helps single-camera associations. The resulting graphs can be clustered using a simple multicut~\cite{chopra1993partition,bansal2004correlation} procedure. 
Unlike existing trackers based on multicuts (\eg~\cite{ristanicvpr2018, nguyen2022lmgp}), our method is entirely online and does not require pre-extracting local tracklets or modeling long-term track relationships. On the contrary, our method works directly on the object detections and provides a sparse representation of all tracks in the form of cross-camera clusters. In particular, our method generalizes well to unseen camera topologies, works with off-the-shelf detection and re-identification networks, and does not require additional annotated information. 

We test our method on two benchmark datasets. On the CityFlow dataset~\cite{tang2019cityflow} (the community standard), we outperform existing online approaches by 14\% in terms of IDF1 (the standard multi-camera tracking metric~\cite{ristani2016MTMC}). On the Synthehicle dataset~\cite{herzog2023synthehicle}, we outperform the existing baseline by more than 25\%. Although designed as a tracking method on the respective image planes, we even show that our method can outperform early-fusion ground plane trackers in cross-scene setups.

%% file: content/02_related.tex
\section{Related Work}
\label{sec:related}
MTMC trackers usually follow the paradigm of
tracking-by-detection~\cite{ristanicvpr2018, hou2019locality, hsu2019multi, luna2022graph, luna2021online, quach2021dyglip, shim2023fast, xu2017cross, he2020multi,  cheng2023rest, nguyen2022lmgp}. Methods differ in many categorical aspects: Some are specifically designed for camera systems with overlapping fields of view. Others focus on long-term tracking and re-identifying targets where cameras are non-overlapping. However, a typical process usually consists of object detection, extraction of appearance features, and subsequent assignment of detections to distinct clusters. In that way, it is closely related to the research area of single-camera MOT~\cite{dendorfer2021motchallenge}, and paradigm-defining methods such as DeepSORT~\cite{Wojke2018deep} have influenced the development of MTMC methods~\cite{naphade20204th}.

Like single-camera MOT, it is usually modeled as a graph problem, where the nodes consist of detections or single-camera tracks, and the edges describe a possible assignment between individual detections or tracks~\cite{hofmanncvpr2013, ristani2016MTMC, nguyen2022lmgp}. The final association between individual measurements is generally obtained by applying a discrete solver, such as min-cost max-flow~\cite{hofmanncvpr2013}, (lifted) multicut~\cite{ristanicvpr2018, nguyen2022lmgp}, k-shortest paths~~\cite{chavdarova2018wildtrack}, Hungarian algorithm~\cite{Teepe_2024_CVPR}, or a hierarchical clustering algorithm~\cite{luna2021online}.
The research community has traditionally concentrated on person tracking for surveillance applications. However, discontinuation of the DukeMTMC dataset~\cite{ristani2016MTMC} for privacy reasons left the community with a sparse selection of short-range and simplistic datasets, causing a shift toward multiple-camera vehicle tracking, particularly within the AICityChallenge \cite{he2019multi, hou2019locality, hsu2019multi, luna2021online, luna2022graph, yang2022box, specker2022improving, shim2023fast}.

The leading paradigm in MTMC is to extract local tracks with a single-camera tracking module and then clustering the local tracks into cross-camera trajectories~\cite{ristanicvpr2018, he2019multi, qian2020electricity, he2020multi, quach2021dyglip, specker2022improving, shim2023fast}. \citet{ristanicvpr2018} suggest a hierarchical method based on correlation clustering that first groups detections into local tracklets. Because they address non-overlapping camera systems, leveraging the accuracy of a single camera reduces the tracking problem to one of person re-identification. \citet{he2019multi} apply hierarchical clustering on a similarity matrix obtained from the trajectory appearance features of single-camera tracks. In~\cite{qian2020electricity}, single-camera tracking is followed by a multi-camera assignment based on the query-gallery-setup of a vehicle re-identification problem. The TRACTA algorithm~\cite{he2020multi} applies restricted non-negative matrix factorization to directly assign tracks to targets in a slight deviation from the typical tracking-by-detection setup. \citet{specker2022improving} suggest a range of single camera track refinements before merging tracks by hierarchical clustering. \citet{nguyen2022lmgp} extract single-camera tracks and associate them in a lifted multicut assignment.
The inherent graph structure of the tracking problem has inspired approaches to learn the association with graph neural networks~\cite{braso2020learning, quach2021dyglip, luna2022graph, cheng2023rest}. DyGLIP~\cite{quach2021dyglip} learns to predict online links of single-camera tracks based on appearance cues. 
\citet{luna2022graph} introduce a global graph neural network where nodes represent single-camera trajectories. \citet{cheng2023rest} introduce a spatial-temporal method based on message-passing networks and associate detections spatially first. Finally, Teepe~\etal aggregate detections spatially and solves the tracking problem in bird eye view with appearance~\cite{teepe2023earlybird} or motion cues~\cite{Teepe_2024_CVPR}.

%% file: content/03_method.tex
\section{Method}
Let $V_t = \{V_t^{(1)}, \dots, V_t^{(M)}\}$ be the set of online video streams for cameras $1, \dots, M$, 
where each video $V_t^{(m)}$ is a time-ordered sequence $\left( \bm I_{t'}^{(m)} \right)_{t' \leq t}$ of images up to frame $t$. We assume the cameras are static, share at least some overlapping field-of-view, and are time-synchronized and calibrated., \ie,
\begin{equation}
\label{eq:video}
    V_t^{(m)} = \left( \left( \bm I_{t'}^{(m)} \right)_{t' \leq t},  \bm P^{(m)} \right), \,\, m=1,\dots,M,
\end{equation}
where, for all m, $\bm I_{t}^{(m)} \in \mathbb R^{W \times H \times C}$, and $\bm P^{(m)} \in \mathbb R^{3x3}$ is a projector from 2D pixel to BEV world coordinates (\eg~a homography matrix).
Our goal is to estimate the ground truth set of multiple-camera tracks, each of which is the cross-camera time-ordered sequence of bounding-boxes that belong to a distinct object identity.

We follow previous work in single- and multiple-camera tracking and formulate tracking as a graph problem~\cite{nguyen2022lmgp}. The tracking is done iteratively as follows:
\begin{enumerate*}
[label=(\roman*)]
    \item Single-camera detections and existing tracks are embedded into a common node representation and weights are calculated between them;
    \item the overall graph structure is adapted for passing it into a multi-cut solver, which returns updated spatial-temporal clusters in a single solver step;
    \item existing tracks are updated and new ones are created based on the clustering results, with some minor additional tricks to make the assignment stable.
\end{enumerate*}
\Cref{fig:teaser} provides a high-level overview of our method, and this section describes the entire process in detail.

\paragraph{Graph formulation} Given the video sequence $\{V_t^m\}_m$, we run an object detector and an appearance feature extractor to obtain detections $\mathcal \{D_t^m\}_m$ and features $\{\mathcal F_t^m\}_m$, respectively, as is common in multi-object tracking~\cite{Wojke2018deep}. Our goal is to find a partition of bounding boxes, where each partition $B_k \in \mathbb R^{M \times F}$ is the collection of cross-camera detections that belong to the same target object, and $F$ is the dimension of the feature space. To this end, we construct a weighted undirected graph $G_t = (\mathcal V_t, \mathcal E_t, w_t)$, where $t$ is the current (fixed) time frame. For each bounding box we define a \emph{measurement}
\begin{align}
\label{eq:superbox}
    b_k = \left( \operatorname{feat}(b_k),  \operatorname{pos_{2D}}(b_k), \operatorname{pos_{BEV}}(b_k) \right) \in \mathbb R^F,
\end{align}
where $\operatorname{feat}(b_k)$ is the appearance feature of $b_k$ (obtained by a CNN), $\operatorname{pos}_{\text{2D}}(b_k)$ is the 2D bounding box, and $\operatorname{pos}_{\text{BEV}}(b_k)$ is the estimated position of the bounding boxes on the ground plane (obtained by applying a projection matrix, \eg~a homography projection). From the set of all $b_k$, we aim to find a partition of detections into cross-camera clusters $\mathcal B_t = \{B_k\}_k$, where for each $B_k \in \mathcal B_t$, two detections $b_i$ and $b_j$ are elements of $B_k$ if and only if they share the same object identity. Thus, each $B_k$ is of the form
\begin{align}
    B_k = \left(b_{k_m} \right)_{1\leq m\leq M}\, \big| \, \operatorname{cam}(b_{k_m}) \neq \operatorname{cam}(b_{k_{m'}})
\end{align}
for all $m \neq m'$, where $b_{k_m}$ is the node in camera $m$ that belongs to $B_k$. To construct $\mathcal B_t$, we add all nodes $\{b_k\}_k$ of the current time frame to $\mathcal V_t$ and add a corresponding edge $(b_i, b_j)$ to $\mathcal E_t$. 
Since we want to solve a tracking problem, and not just a detection problem, the cross-view clusters need to be connected to existing tracks (which are also cross-view clusters). Suppose for now that $\mathcal B_t$ is already constructed, we also want to connect its elements $B_k$ to the clusters from the previous time step $t-1$. This conforms to constructing sequences of clusters
\begin{align}
    S_k = \left( B_{k_{t'}} \right)_{t' \leq t}.
\end{align}
Each $B_k$ is a cross-view detection belonging to a target identity and $S_k$ is the identity-consistent temporal collection of such detections.

In the following sections, we refer to $B_k$ as \emph{superbox} because it is essentially a superset of individual detections. A core idea of our approach is to define the tracks $S_k$ in a way that they are represented by a \emph{single} superbox $B_k \in \mathbb R^{M \times F}$, respectively, by temporal averaging and a corresponding assignment procedure, enabling similarity computation between single detections and entire spatial-temporal clusters. It remains to show how to determine the connectivity and weights between \begin{enumerate*}[label=(\roman*)]
  \item two individual detections
  \item a superbox and a detection
  \item and between superbox and existing tracks
\end{enumerate*}.

\paragraph{Weight computation} Our goal is to find a suitable way to compare individual single-camera detections with multi-camera detections and spatial-temporal tracks. Recall that every multi-camera detection is a spatial-temporal track with only one time-step, and every individual single-camera detection is a multi-camera detection with only one available camera. 
Based on this, we define the feature of a superbox as the vector of its node features, \ie,
\begin{align}
    \feat\left( B_k \right) = \left( \feat\left(b_m^k\right)\right)_{m=1}^M,
\end{align}
and reset the feature of an individual node as
\begin{align}
    \feat\left(b_m^k\right) \leftarrow \begin{cases}
        \feat(b_m^k) & \mbox{ if } b_m^k \neq \varepsilon \\
        \nanmean\limits_{m' \neq m} \feat(b_{m'}) & \mbox{ else.}
    \end{cases}
\end{align}
That is, in cameras where there is no evidence for a superbox (\eg~due to occlusion or non overlapping areas), the evidence is \emph{replaced} by aggregation of available evidence. Finally, for a track, the evidence is aggregated over time:
\begin{align}
    \feat\left( S_k \right)_m = \EMA_{t' \leq t} \feat\left( B_k^{t'}\right)_m , && m\in [M],
\end{align}
where $\EMA$ denotes an exponential moving average. The values $\operatorname{pos_{2D}}$ and $\operatorname{pos_{BEV}}$ of a superbox and tracks are reset analogously. Note that by using aggregations, each spatial-temporal track has a strong but \emph{sparse} representation in the appearance-position space, since $S_k$ is now represented by a single superbox, \ie, we have $S_k \in \mathbb R^{M \times F}$. The weight between two tracks is then defined as the convex combination 
\begin{align}
\label{eq:scale}
    w_{i, j} = \lambda \feat\left(S_i , S_j\right) + (1- \lambda) \pos\left(S_i , S_j\right),
\end{align}
where $\feat\left(S_i, S_j\right)$ is the (scaled) cosine feature similarity between $S_i$ and $S_j$ and $\pos\left(S_i , S_j\right)$ is the (scaled) positional similarity between $S_i$ and $S_j$. The latter is derived from the Euclidean distance by inversion and thresholding, and the former is a rescaled similarity in which similar nodes should have a value close to $1$, and dissimilar nodes have a value close to $-1$ (see implementation details). With $w_{i,j}$ defined this way, we can now compare individual detections with entire spatial-temporal tracks, allowing subsequent solving in a single spatial-temporal step.

\paragraph{Constraints} Before passing the weights to the solver, we need to ensure the tracking constraints. To this end, we define a set of infeasible edges $\mathcal E_{\inf}$. In it are all edges $(i, j)$ for which $S_i$ and $S_j$ have evidence at the same time and frame, \ie,
\begin{align}
    \mathcal E_{\inf_1} = \{ &(i, j) \in E_t \,|\, \exists t' \leq t, m \leq M : \\ 
    & (S^i_m)_{t'}, (S^j_m)_{t'}  \neq \varepsilon\}.
\end{align}
Additionally, we mark edges $(i, j)$ as infeasible whenever the ground plane distance between $S_i$ and $S_j$ exceeds the distance threshold, \ie,
\begin{align}
    \mathcal E_{\inf_2} = \{ &(i, j) \in E_t \,|\, \operatorname{pos}\left( S_i, S_j \right) > \delta_{\text{pos}} \}.
\end{align}
We prune the set of infeasible edges $\mathcal E_{\inf} = \mathcal E_{\inf_1} \cup \mathcal E_{\inf_2}$ by reassigning a large negative penalty weight $\rho \ll 0$ to them \ie, $w_{i, j} = \rho, \forall (i, j) \in  \mathcal E_{\inf}$, as is common in the multi-cut literature~\cite{nguyen2022lmgp}.

\paragraph{Matching} We use the minimum-cost multicut formulation~\cite{chopra1993partition, bansal2004correlation, beier2016efficient, keuper2015efficient, abbas2022rama} to compute clusters of nodes that belong to the same identity. In this formulation, a negative weighted edge $e$ between two nodes ($w_e < 0$) indicates that the two nodes repel each other, while a positive weight ($w_e > 0$) indicates that the two nodes attract each other. Formally, for a weighted, undirected graph $G=(\mathcal V, \mathcal E, w)$, the minimum-cost multicut problem 
\begin{mini!}%
{\strut y \in \{0, 1\}^{|E|}}%
 { \sum_{e \in E} w_e y_e \label{eq:mcmin}}%
{}%
{}
\addConstraint{y_e}{\leq \sum_{e' \in C \setminus \{ e \} } y_{e'}\label{eq:mcconstr1}}
\addConstraint{}{\forall C \in \cycles(G) \; \forall e \in C\label{eq:mcconstr2}}
\end{mini!}
aims to find a cost-optimal edge labeling $y \in \{0, 1\}^{|E|}$ that partitions the nodes into disjoint clusters. 
The cycle-constraint ensures that nodes within a cluster cannot be connected to nodes in other clusters. The problem is also known as \textit{correlation clustering}~\cite{bansal2004correlation} and was used as such by Ristani et al. \cite{ristanicvpr2018}.

\paragraph{Assignment and Track Management} From the edge labeling $y \in \{0, 1\}^{|E|}$ we obtain a set of $L \leq \mathcal |V|$ pairwise disjoint clusters $\{C_{\ell}\}_{\ell}$. We sort the nodes of each cluster by time and assign the identities as follows. Let $C_{\ell}$ be the $\ell$-th cluster and let $t$ be the current frame.
\begin{itemize}
    \itemsep -1pt
    \item If all nodes are from the current time step $t$, none of the nodes in $C_{\ell}$ have an object identity. We assign a new unused object identity to the entire cluster and start it as new track.
    \item Otherwise, select the newest node $p \in C_{\ell}$ such that $\operatorname{time}(p) < t$ as pivot element. We aggregate features and positions spatially among all cluster nodes from time step $t$, then update the features and positions of $p$ using moving-average aggregation as described. Continue to track $p$ as representative of the entire cluster.
    \item Unmatched tracks are deactivated. Tracks that are unmatched for longer than the value \emph{patience} are declared lost (as is common in single-camera tracking~\cite{seidenschwarz2023simple}). Lost tracks are subject to similarity decay and considered for matching but without distance thresholding.
    \item A lost track that is unmatched for longer than the value \emph{memory} is killed.
\end{itemize}

After each time-step, we project the existing tracks into the next time-step using the simple linear motion model
\begin{align}
    \widehat{\pos}_{\text{BEV}}\left( b_k^t \right) &\approx \pos_{\text{BEV}}\left( b_k^{t-1} \right) + \operatorname{velo}_{\text{BEV}}\left( b_k^{t-1} \right), \\
    \widehat{\pos}_{\text{2D}}\left( b_k^t \right) &\approx \pos_{\text{2D}}\left( b_k^{t-1} \right) + \operatorname{velo}_{\text{2D}}\left( b_k^{t-1} \right),
\end{align}
where $\pos_{\text{BEV}}$, $\pos_{\text{2D}}$, $\operatorname{velo}_{\text{BEV}}$ and $\operatorname{velo}_{\text{2D}}$ denote the BEV and 2D positions and velocities, respectively. This approach differs from the majority of works which use a Kalman filter as motion model. However, \cite{seidenschwarz2023simple} noticed that a simple linear model is often sufficient for multiple object tracking and we find that a linear motion model is easier to manage within our method. 

\paragraph{Similarity decay}
For a lost track $S_{\ell}$ we \emph{decay} its feature similarity over time. If the track is lost for $\Delta t$ frames, the similarity weight is assigned to $w_{\ell, k} \leftarrow \beta^{\Delta t} w_{\ell, k}$, where $\beta \in (0, 1)$ is a decay factor. The procedure ensures that matching a long-lost track with a current track or detection needs ever stronger appearance cues. This might be counter-intuitive at first, since in single-camera person tracking, the matching conditions between lost and existing tracks are usually relaxed rather than tightened. However, note that vehicles have a lower appearance variance than people and that we lose valuable positional information when tracks are lost. Because we rely on appearance information alone when matching lost tracks, we find that decaying the feature similarity to existing tracks is beneficial when dealing with noisy calibrations and asynchronous cameras.

\paragraph{IoU-Matching} 
Finally, we want to utilize the \emph{intersections over union} (IoU) of bounding boxes within single cameras, which are the basis of all single camera tracking methodologies. As opposed to appearance and ground plane positional information, which can be easily aggregated into strong multi-camera cues, the IoU is bound to single cameras. We solve this problem with a simple procedure: Instead of using the IoU directly, we add an IoU bias value $b_{\text{IoU}}$ to the edge weight of an existing track and a detection if and only if the detection is the Hungarian match between that track and all other available detections. Optionally, we prune all other edges between a detection and a track by setting its weight to zero. Our IoU-matching strategy differs from the pre-clustering strategy suggested in~\cite{nguyen2022lmgp} which works on nearest neighbors on the BEV ground plane, rather than the individual image planes. Note that the Hungarian algorithm is only used to decide on which edge weights the IoU bias should be added to. The actual matching is done entirely by the multicut solver in a single step.

%% file: content/04_results.tex
\section{Results}
\paragraph{Implementation details} For appearance feature extraction, we use a ResNet-50~\cite{DBLP:conf/cvpr/HeZRS16} LightMBN~\cite{herzog2021lightweight} network for Synthehicle and a ResNet-101 LCFractal~\cite{liu2021city} network for CityFlow, loading the pre-trained weights provided by the authors. Both networks were trained on the train splits of the respective datasets. For object detection, we use YOLOX-X~\cite{yolox2021} pre-trained on COCO~\cite{lin2014microsoft} and fine-tuned on the training data for Synthehicle, and without further fine-tuning for CityFlow. We use RAMA~\cite{abbas2022rama} to solve multicut instances fast and directly on the GPU. The penalty for infeasible weights is set to $\rho = -100$. We noticed numerical instabilities for significantly smaller values. The LAP solver used for Hungarian pre-matching is from scipy~\cite{2020SciPy-NMeth}. To compute ground plane positions of the tracks, we project the 2D image plane bounding boxes using the homography matrix of the respective camera. Specifically, for a bounding box $\mathbf b = (t, l, w, h)$ and homography matrix $\mathbf H$, we compute $(x=\frac{\tilde x}{\tilde z}, y=\frac{\tilde y}{\tilde z})$ with
\begin{align}
\label{eq:projection}
    (\tilde x, \tilde y, \tilde z) = \mathbf H \left(
    l + \frac{1}{2}w,
    t + {\alpha} h,
    1 \right)^T,
\end{align}
where ${\alpha} \in [0, 1]$ determines the amount of height added to the bounding box center width in the projection. We find that ${\alpha}=0.85$ produces the best results, which corresponds to taking \emph{almost} the centers of the bottom of the bounding box as a basis for projection. The weights (cf. \Cref{eq:scale}) are scaled as follows: We linearly rescale the appearance features so that the appearance similarities are mapped to $[-1, 0)$ if they are smaller than a rescale threshold $\theta_{\text{feat}}$, and to $(0, 1]$ otherwise. Distances are converted to positional similarity using a distance threshold $\theta_{\text{pos}}$, \ie, $\text{pos}(S_i,S_j) = 1 - \text{dist}(S_i,S_j)/\theta_{\text{pos}}$. 

\paragraph{Datasets} The CityFlow dataset~\cite{tang2019cityflow} has been part of the AICityChallenge for several years and consists of six scenarios and over $300,000$ annotated bounding boxes. Only two scenes, S01 and S02, have overlapping cameras. As is common in the literature~\cite{luna2021online, luna2022graph, shim2023fast}, we evaluate the S02 scene and keep S01 as training data for fine-tuning and hyperparameter search. %
The Synthehicle~\cite{herzog2023synthehicle} dataset contains 64 synthetic scenes with over 4 million annotated bounding boxes for multiple-camera multiple-vehicle tracking. Each of the scenes has been recorded in four different weather scenarios, including rain and night, and consists of three to eight cameras. Half of the scenes are for tracking in overlapping cameras, and we evaluate our tracker on the 12 overlapping test scenes. %
\paragraph{Metrics} We report performance using the IDF1, IDR and IDP metrics introduced in~\cite{ristani2016MTMC}, which essentially measure consistency of identity assignment and have been used traditionally to evaluate multi-camera tracking performance. On CityFlow, we evaluate on the image plane and on Synthehicle, we evaluate on the image and on the ground plane. For the latter, we report the IDF1 value along with the CLEAR \cite{Bernardin:2008:CLE} metric MOTA. Matching distance thresholds are measured using the intersection over union of bounding boxes on the image plane ($\text{IoU}=0.5$) and using Euclidean distance ($r=1$ meters) on the ground plane.

\input{content/tables/results_aicity}
\input{content/tables/results_synthehicle_img}
\input{content/tables/results_synthehicle_bev}

\begin{figure*}[htbp]
    \centering
    \subfloat[Grid search on Synthehicle (train).]{
        \includegraphics[width=0.30\textwidth]{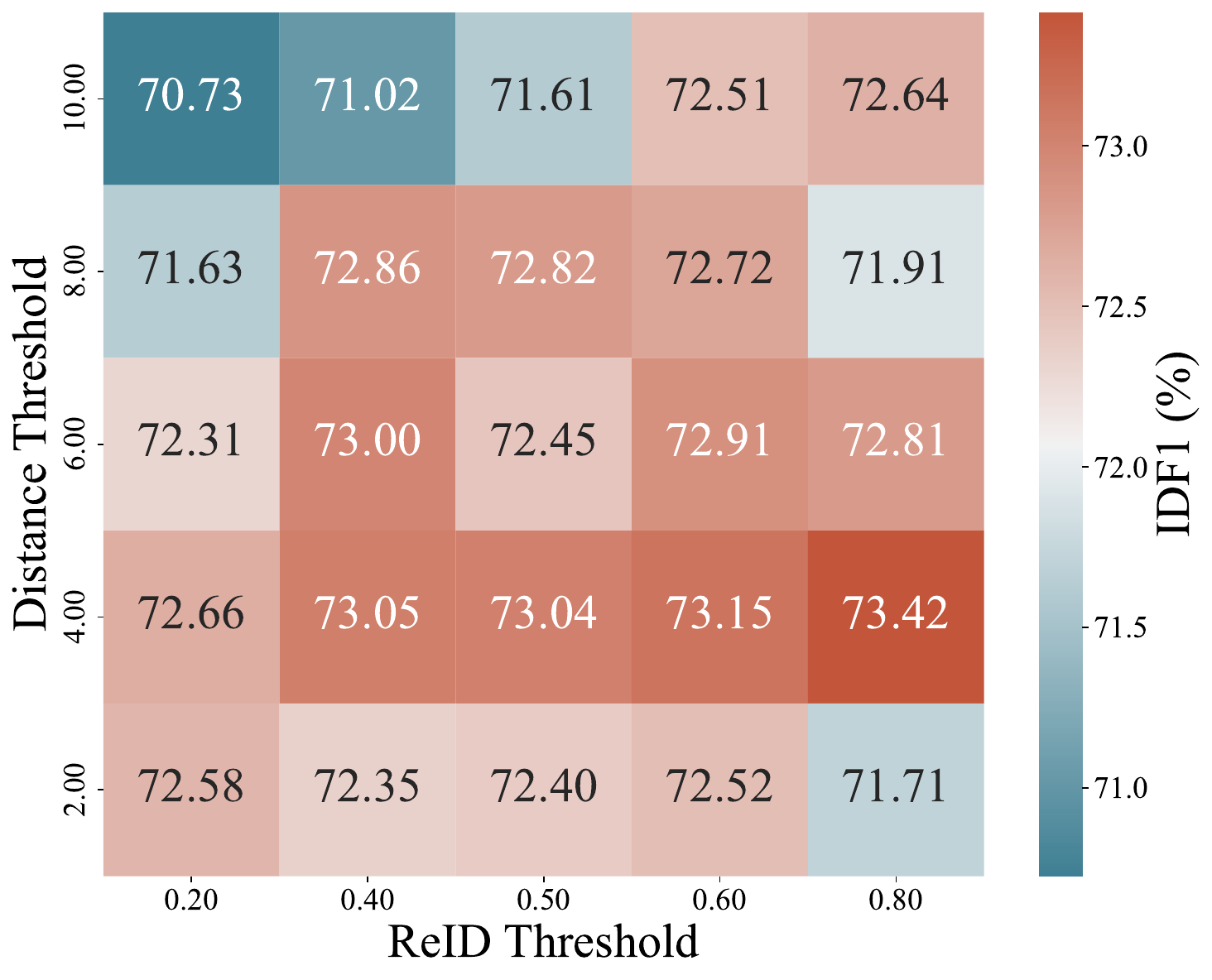}
        \label{fig:figure1}
    }
    \hfill
    \subfloat[Grid search on CityFlow S01.]{
        \includegraphics[width=0.295\textwidth]{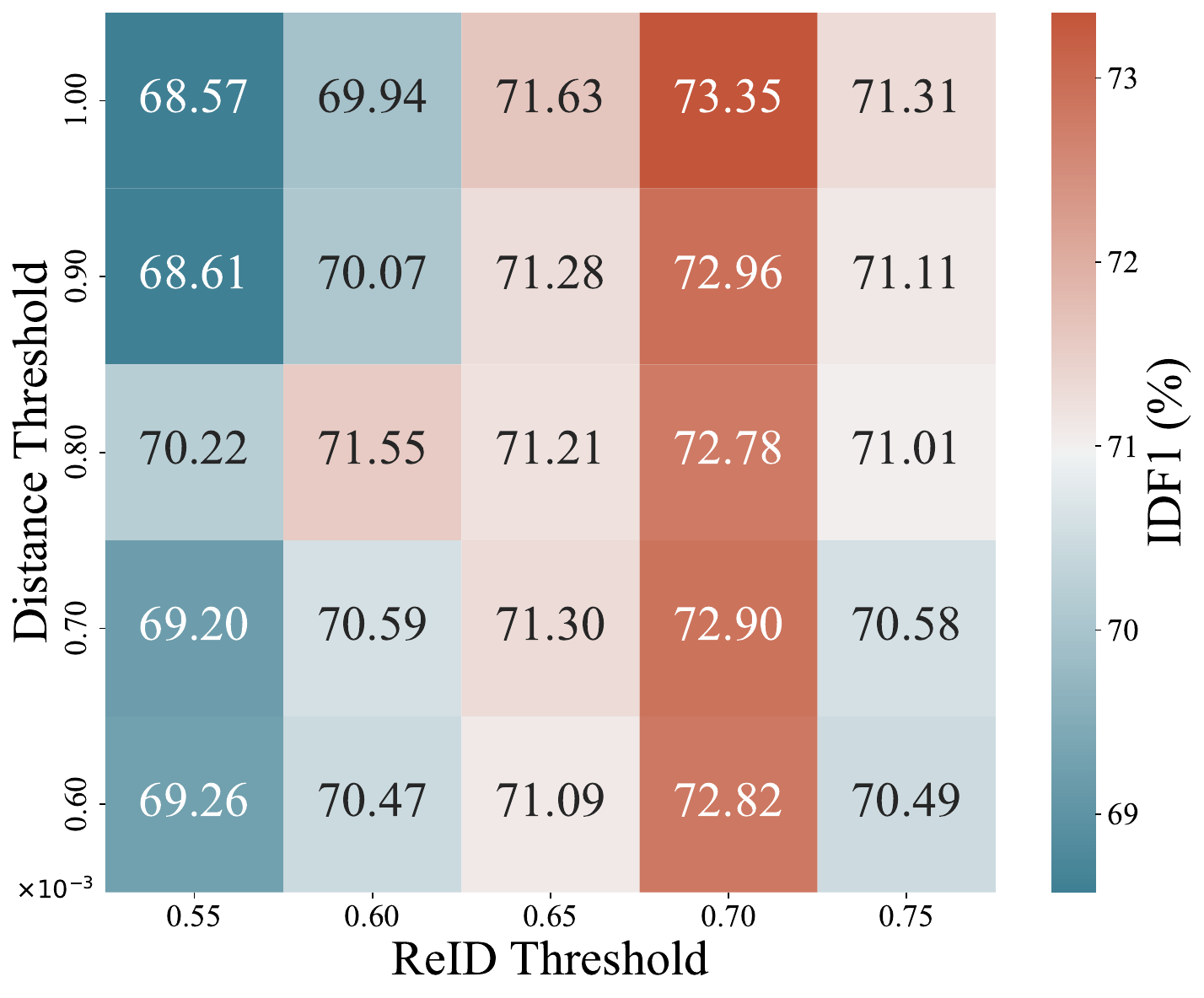}
        \label{fig:figure2}
    }
    \hfill
    \subfloat[Importance of ReID vs position (test).]{
        \includegraphics[width=0.33\textwidth]{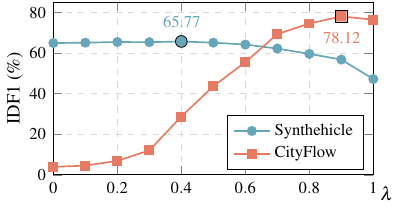}
        \label{fig:figure3}
    }
    \caption{Matching thresholds for ReID and positions and their importance. We perform a grid search on the training subsets of Synthehicle (a) and CityFlow (b) to find the optimal parameters for weight scaling. For $\theta_{\text{feat}}$ and $\theta_{\text{pos}}$ we choose $(0.8, 4.0)$ and $(0.7, 0.001)$ for Synthehicle and CityFlow, respectively. The relative importance of the thresholds is depicted in Figure (c), and we set $\lambda$ (cf. \cref{eq:scale}) to $0.4$ for Synthehicle and to $0.9$ for CityFlow, \ie, for CityFlow. The ground plane is calibrated in meters for Synthehicle and in GPS for CityFlow.}
    \label{fig:allfigures}
\end{figure*}

\paragraph{Benchmark results} \Cref{tab:results_cityflow} lists results on the on the CityFlow dataset, where we outperform other fully-online trackers by more than 14\% in IDF1. In particular, we even outperform all offline methods in IDP. \Cref{tab:tracking_results} lists our results on the Synthehicle dataset. Despite using the same detection weights as the baseline, we outperform this baseline by more than 20\% in IDF1, IDP, and almost in IDR. We believe this is mainly due to our method better utilizing 2D and 3D positional cues.

\paragraph{ReID vs. distance thresholds} \Cref{fig:allfigures} depicts hyperparameter grid searches on Synthehicle and CityFlow to determine the optimal values for weight scaling and their relative importance. We used the respective train splits of the datasets to find the optimal hyperparameters. For $\theta_{\text{feat}}$ and $\theta_{\text{pos}}$ we choose $(0.8, 4.0)$ and $(0.7, 0.001)$ for Synthehicle and CityFlow, respectively. 
\paragraph{ReID vs. distance importance} The importance of appearance feature similarity and distances between two tracks highly depends on the dataset, as can be seen in \Cref{fig:allfigures} (c). Distance values are not reliable for CityFlow, because the provided homography matrix is based on GPS values with high reprojection error, and the cameras are not synchronous (\ie, through time, the GPS coordinates will become almost meaningless for comparison). Here, re-identification features are much more important. The reverse is true for Synthehicle. Because it is a synthetic dataset camera calibration is almost perfect, but many distinct vehicles look \emph{exactly alike}, \ie, they share the \emph{same} appearance features. Here, the distance threshold is much more important than the re-identification features. 

\paragraph{Memory and patience} Reasonable patience values (\ie, the time an inactive track is tracked before being marked as \emph{lost}) are in the range of 0 to 3. The optimal memory value (\ie, the time a lost track is kept before being deleted) depends on  camera synchronicity. For very asynchronous cameras (like in CityFlow), a high value (100-200) is beneficial, whereas for Synthehicle, high values yield lower performance since they increase the likelihood of identity switches. We chose 160 as memory value for CityFlow, and 15 as memory value for Synthehicle.

\input{content/tables/ablation_studies}
\paragraph{Ablation studies} \Cref{tab:ablation} compares the influence of the core components of our method. The linear assignment for IoU-based matching (\emph{pre-matching}) proves essential on the Synthehicle dataset. It successfully adds a single-camera bias in dense scenes with many occlusions, where appearance, motion, and ground plane geometric information are degraded. Pruning edges implies setting spatial-temporal edges outside the linear assignments to zero, and it proves beneficial on Synthehicle.
On CityFlow, IoU matching negatively impacts performance. We assume this is because traffic flow in CityFlow is less dense, and re-identification features remain the most vital cue in all situations. Similarity decay is particularly important for CityFlow, where we keep lost tracks in memory for a high number of frames due to camera asynchronicity. To prevent identity switches, similarity decay ensures that new detections are only assigned to lost when the evidence is strong.

\paragraph{Weaknesses} While our method achieves state-of-the-art performance, we want to underline some drawbacks. First, we obviously cannot compete with methods that train on the target camera geometry. The benefit from this is that the tracker should work \emph{out-of-the-box} across scenes, however, some key hyperparameters still need to be tuned to the target scene (\eg~the matching thresholds). Another drawback lies in our methodology itself: Solving the problem in combined spatial-temporal steps degrades some of the motion and appearance information and some tricks are required to mitigate this (such as the similarity decay and the Hungarian pre-matching). As usual, there is no free lunch.

\paragraph{Ethical implications} While tracking methods can be used to benefit humanity (\eg, by increasing traffic safety or enabling sports analysis), they can be misused for dubious applications (\eg, mass surveillance). Datasets such as DukeMTMC and WildTrack have been critically discussed due to privacy concerns~\cite{Exposing.ai}. Therefore, we welcome the development of synthetic datasets that at least exclude acute privacy concerns, although they can also potentially be used for dubious purposes. Our method focuses on applications in the area of smart cities rather than surveillance, and we hope that it will only be used for those purposes.

%% file: content/tables/results_aicity.tex
\begin{table}
\center
\caption{Performance comparison on CityFlow. Results computed on the S02 validation seen. Our method outperforms all online methods regarding IDF1, and even all offline methods regarding IDP.  \,\cxmark: Applies offline manual time synchronization of cameras.}
\resizebox{\linewidth}{!}{%
\begin{tabular}{ r  r c r r r}
 \toprule
 & Detector  & online & IDF1 & IDP & IDR \\
 \midrule
BUPT~\cite{he2019multi} & FPN~\cite{lin2017feature} &\xmark& 70.22 & 78.23 & 63.69 \\ 
ANU~\cite{hou2019locality} & SSD~\cite{liu2016ssd}&\xmark& 74.06 & 67.53 & 81.99 \\
UWIPL~\cite{hsu2019multi} & Mask R-CNN~\cite{he2017mask} &\xmark& 79.87 & 70.21 & 92.61 \\
GCN~\cite{luna2022graph} & SSD~\cite{liu2016ssd} & \xmark & 81.06 & 71.95 & 92.81 \\
\midrule
Luna et al.~\cite{luna2021online} & Mask R-CNN~\cite{he2017mask} & \cmark & 63.76 & 57.23 & 71.99 \\
Luna et al.~\cite{luna2021online} & EfficientDet~\cite{tan2020efficientdet} & \cmark & 64.26 & 55.15 & 76.98 \\
DyGLIP~\cite{quach2021dyglip} & Mask R-CNN~\cite{he2017mask} & \cmark  & 64.90 & -- & --  \\
Shim et al.~\cite{shim2023fast} & YOLOv7~\cite{wang2023yolov7} & \cxmark & 78.40 & 78.80 & 78.00 \\
\midrule
 \ours & YOLOX-X~\cite{yolox2021} & \cmark & \bftab 79.58 & \bftab 81.10 & \bftab 78.11  \\
 \bottomrule
\end{tabular}}
\label{tab:results_cityflow}
\end{table}

%% file: content/tables/results_synthehicle_img.tex
\begin{table}[htbp]
\centering
\caption{Performance comparison on Synthehicle (image plane). Results computed for the overlapping test scenes only. We use the YOLOX-X detector with the same weights as the offline baseline. All values in percent. Best values printed bold.}
\resizebox{0.92\linewidth}{!}{%
\begin{tabular}{@{}l*{6}{c}@{}}
\toprule
& \multicolumn{3}{c}{Baseline~\cite{herzog2023synthehicle}} & \multicolumn{3}{c}{Ours} \\
\cmidrule(lr){2-4} \cmidrule(lr){5-7}
Scene &  \small{IDF1} & \small{IDP} &  \small{IDR} & \small{IDF1} &  \small{IDP} &  \small{IDR} \\
\midrule
Town06-O-day & 46.7 & 55.3 & 40.4 & \bftab 82.2 & \bftab 85.3 & \bftab 81.5 \\
Town06-O-dawn & 45.4 & 55.0 & 38.7 & \bftab 75.6 & \bftab 79.9 & \bftab 76.1 \\
Town06-O-rain & 45.9 & 60.5 & 37.8 & \bftab 73.4 & \bftab 80.0 & \bftab 67.7 \\
Town06-O-night & 50.1 & 58.8 & 43.7 & \bftab 74.3 & \bftab 75.6 & \bftab 73.1 \\
\midrule 
Town07-O-day & 37.1 & 41.0 & 33.8 & \bftab 70.4 & \bftab 81.8 & \bftab 68.6 \\
Town07-O-dawn & 39.6 & 43.4 & 36.8 & \bftab 71.9 & \bftab 83.4 & \bftab 76.5 \\
Town07-O-rain & 42.6 & 49.1 & 38.3 & \bftab 61.6 & \bftab 72.4 & \bftab 57.9 \\
Town07-O-night & 30.3 & 32.9 & 28.3 & \bftab 49.1 & \bftab 72.4 & \bftab 41.7 \\
\midrule 
Town10-O-day & 25.8 & 27.1 & 24.6 & \bftab 69.1 & \bftab 87.2 & \bftab 69.8 \\
Town10-O-dawn & 31.9 & 34.9 & 30.0 & \bftab 63.4 & \bftab 71.8 & \bftab 60.6 \\
Town10-O-rain & 30.9 & 36.1 & 27.8 & \bftab 51.7 & \bftab 68.1 & \bftab 46.5 \\
Town10-O-night & 24.5 & 30.7 & 23.3 & \bftab 49.6 & \bftab 61.7 & \bftab 45.1 \\
\midrule 
Total & 37.5 & 43.7 & 33.6 & \bftab 66.0 & \bftab 76.6 & \bftab 63.8 \\
\bottomrule
\end{tabular}}
\label{tab:tracking_results}
\end{table}

%% file: content/tables/results_synthehicle_bev.tex
\begin{table}
\center
\caption{Performance comparison on Synthehicle (ground plane). Results computed for the overlapping test scenes only. A same-scene split implies the method has been trained on parts of the same scene as it is tested on, which we don't do.}
\resizebox{\linewidth}{!}{%
\begin{tabular}{ r c c r r}
 \toprule
 & Lifting Method & Split & IDF1 & MOTA\\
 \midrule
 {\color{gray} \small{TrackTacular}~\cite{Teepe_2024_CVPR}} & {\color{gray} Bilinear} {\color{gray} Sampling} & {\color{gray} same} & {\color{gray} 48.0} & {\color{gray}10.6} \\ 
{\color{gray} \small{TrackTacular}~\cite{Teepe_2024_CVPR}} & {\color{gray} Depth Splatting} & {\color{gray}same} & {\color{gray}57.2} & {\color{gray}33.1} \\ 
\midrule
\small{TrackTacular}~\cite{Teepe_2024_CVPR} & Bilinear Sampling & cross & 18.3 & -22.0 \\ 
\small{TrackTacular}~\cite{Teepe_2024_CVPR} & Depth Splatting & cross & 24.2 & \underline{-1.5} \\ 
 \midrule
  \ours  & \Cref{eq:projection}, ${\alpha}=1.00$ & cross & 25.0 & -30.9 \\
  \ours  & \Cref{eq:projection}, ${\alpha}=0.85$ & cross & \bftab 39.4 & \bftab {-0.01}\\
 \bottomrule
\end{tabular}}
\label{tab:results_synthehicl2}
\end{table}

%% file: content/tables/ablation_studies.tex
\begin{table}
\centering
\caption{Ablation studies: We compare the influence of adding key components of our method on the CityFlow and Synthehicle datasets. Evaluation is done on image plane. All values in percent.}
\label{tab:ablation}
\resizebox{0.9\linewidth}{!}{%
\begin{tabular}{lcccccc}
\toprule
 & \multicolumn{6}{c}{Ablation Configurations} \\
\cmidrule(lr){2-7}
 & (1) & (2) & (3) & (4) & (5) & (6) \\
\midrule
Similarity decay & \xmark & \cmark & \xmark & \cmark & \xmark & \cmark \\
Pre-matching & \xmark & \xmark & \cmark & \cmark & \cmark & \cmark \\
Pruning & \xmark & \xmark & \xmark & \xmark & \cmark & \cmark \\
\midrule
IDF1 (CityFlow) & 74.6 & \bftab 79.5 & 74.2 & 78.1 & 69.3 & 69.5 \\
IDF1 (Synthehicle) & 58.1 & 59.0 & 60.9 & 61.1 & \bftab 66.0 & 65.7 \\
\bottomrule
\end{tabular}}
\end{table}

%% file: content/05_conclusion.tex
\section{Conclusion} We introduce a new method for multi-target multi-camera vehicle tracking in online scenarios based on multi-cuts. Unlike previous methods, we solve the problem across views and frames iteratively in combined spatial-temporal steps. Our underlying graph formulation enables the efficient and effective comparison of sparse embedded single-camera observations with entire multi-camera tracks. A simple track management combined with intuitive tricks, like similarity decay and IoU-based pre-matching, suffices to outperform all other online methods on the CityFlow and Synthehicle datasets.